\def\year{2020}\relax
\documentclass[letterpaper]{article} 
\usepackage{aaai20}  
\usepackage{times}  
\usepackage{helvet} 
\usepackage{courier}  
\usepackage[hyphens]{url}  
\usepackage{graphicx} 
\usepackage{graphicx}  
\usepackage{amsfonts}
\usepackage{amsmath}
\usepackage{amsthm}
\usepackage{booktabs}
\usepackage{bm}
\usepackage{subcaption}
\usepackage{tablefootnote}
\usepackage[title]{appendix}
\usepackage{wrapfig}
\title{Hyperbolic Interaction Model for Hierarchical Multi-Label Classification}
\author{Boli Chen, Xin Huang, Lin Xiao, Zixin Cai, Liping Jing\\ 
Beijing Jiaotong University, Beijing, China\\ 
\{18120345, 18120367, 17112079, 18120340, lpjing\}@bjtu.edu.cn 
}

\begin{document}

\maketitle

\begin{abstract}
  Different from the traditional classification tasks which assume mutual exclusion of labels, hierarchical multi-label classification (\textit{HMLC}) aims to assign multiple labels to every instance with the labels organized under hierarchical relations. Besides the labels, since linguistic ontologies are intrinsic hierarchies, the conceptual relations between words can also form hierarchical structures. Thus it can be a challenge to learn mappings from word hierarchies to label hierarchies. We propose to model the word and label hierarchies by embedding them jointly in the hyperbolic space. The main reason is that the tree-likeness of the hyperbolic space matches the complexity of symbolic data with hierarchical structures. A new Hyperbolic Interaction Model (\textit{HyperIM}) is designed to learn the label-aware document representations and make predictions for \textit{HMLC}. Extensive experiments are conducted on three benchmark datasets. The results have demonstrated that the new model can realistically capture the complex data structures and further improve the performance for \textit{HMLC} comparing with the state-of-the-art methods. To facilitate future research, our code is publicly available.
\end{abstract}

\section{Introduction}

Traditional classification methods suppose the labels are mutually exclusive, whereas for hierarchical classification, labels are not disjointed but organized under a hierarchical structure. Such structure can be a tree or a \textit{Directed Acyclic Graph}, which indicates the parent-child relations between labels. Typical hierarchical classification tasks include protein function prediction in bioinformatics tasks \cite{wehrmann2017hierarchical}, image annotation \cite{dimitrovski2011hierarchical} and text classification \cite{meng2019weakly}. In this paper, we focus on hierarchical multi-label text classification, which aims to assign multiple labels to every document instance with the labels hierarchically structured.

In multi-label classification (\textit{MLC}), there usually exist a lot of infrequently occurring \textit{tail labels} \cite{bhatia2015sparse}, especially when the label sets are large. The fact that \textit{tail labels} lack of training instances makes it hard to train an efficacious classifier. Fortunately, the effectiveness of utilizing label correlations to address this problem has lately been demonstrated. In literatures, label correlations can be determined from label matrix \cite{zhang2018deep} or label content \cite{wang2018joint}. The main idea is to project the labels into a latent vectorial space, where each label is represented as a dense low-dimensional vector, so that the label correlations can be characterized in this latent space. For hierarchical multi-label classification (\textit{HMLC}), labels are organized into a hierarchy and located at different hierarchical levels accordingly. Since a parent label generally has several child labels, the number of labels grows exponentially in child levels. In some special cases, most labels are located at the lower levels, and few training instances belong to each of them. In other words, \textit{tail labels} also exist in \textit{HMLC}. Different from the traditional \textit{MLC}, the label structure, which is intuitively useful to detect label correlations, is well provided in \textit{HMLC}. 

\begin{figure*}
  \centering
  \begin{subfigure}{0.38\linewidth}
    \centering
    \includegraphics[width=.61\linewidth]{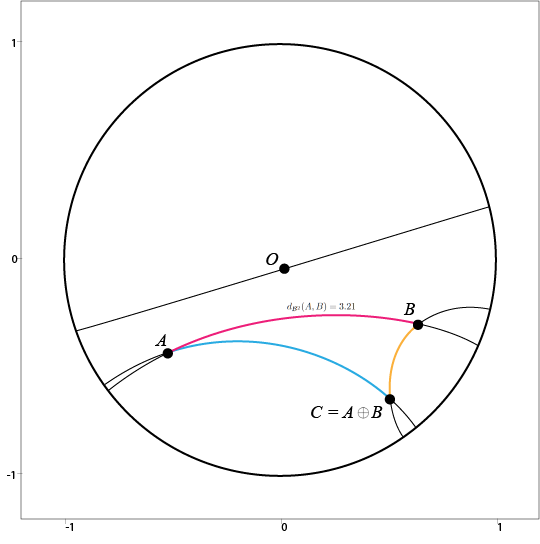}
    \caption{Visualization of geodesics and M\"{o}bius addition}
    \label{fig:mob_add}
  \end{subfigure}
  \begin{subfigure}{0.38\linewidth}
    \centering
    \includegraphics[width=.61\linewidth]{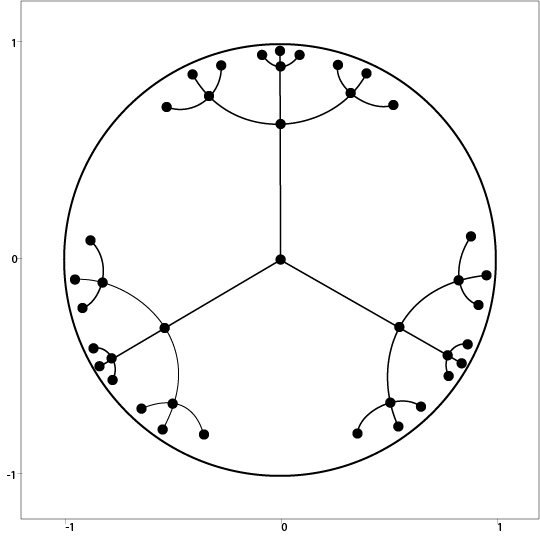}
    \caption{A tree embedded in the Poincaré disk}
    \label{fig:embed_tree}
  \end{subfigure}
  \caption{(a) Point $ C $ represents the M\"{o}bius addition of point $ A $ and $ B $. In the Poincaré disk model, geodesics between points are arcs and perpendicular to its boundary due to its negative curvature. (b) The line segments indicate the geodesics between each pair of connected nodes in a tree.}
  \label{fig:poincare}
\end{figure*}

Inspired by recent works on learning hierarchical representations \cite{nickel2017poincare}, we propose to embed the label hierarchy in the hyperbolic space. Taking advantage of the hyperbolic representation capability, we design a Hyperbolic Interaction Model (\textit{{HyperIM}}) to classify hierarchically structured labels. \textit{{HyperIM}} embeds both document words  and labels jointly in the hyperbolic space to preserve their latent structures (\textit{e.g.} structures of conceptual relations between words and parent-child relations between labels). Semantic connections between words and labels can be furthermore explicitly measured according to the word and label embeddings, which benefits extracting the most related components from documents and constructing the label-aware document representations. The prediction is directly optimized by minimizing the \textit{cross-entropy} loss. Our contributions are summarized as follows:

\begin{itemize}
  \item We adopt hyperbolic space to improve \textit{HMLC}. A novel model \textit{{HyperIM}} is designed to embed the label hierarchy and the document text in the same hyperbolic space. For the classification, semantic connections between words and labels are explicitly measured to construct the label-aware document representations.
  \item We present \textit{partial interaction} to improve the scalability of the interaction model. For large label spaces, negative sampling is used to reduce the memory usage during interaction.
  \item Extensive experiments on three benchmark datasets show the effectiveness of \textit{HyperIM}. An ablation test is performed to demonstrate the superiority of the hyperbolic space over the Euclidean space for \textit{HMLC}. In addition, our code is publicly available.
\end{itemize}

\section{Preliminaries}
\label{sec2}

Let $ \mathcal{X} $ denote the document instance space, and let $ \mathcal{L} = \{l_i\}_{i=1}^C $ denote the finite set of $ C $ labels. Labels are organized under a hierarchical structure in \textit{HMLC}, $ \mathcal{T} = \{(l_p, l_q) \ | \ l_p \succeq l_q, \ \  l_p, l_q \in \mathcal{L} \} $ denotes their parent-child relations, where $ l_p $ is the parent of $ l_q $. Given the text sequence of a document instance $ \bm{x} \in \mathcal{X} $ and its one-hot ground truth label vector  $ \bm{y} \in \{0, 1\}^C $, the classification model learns the document-label similarities, \textit{i.e.} the probabilities for all the labels given the document. Let $ \bm{p} \in [0, \ 1]^C $ denote the label probability vector predicted by the model for $ \bm{x} $, where $ \bm{p}_{[i]} = P( l_i \ | \ \bm{x} ) $ for $ l_i \in \mathcal{L} \ (i = 1, \dots, c) $ (the subscript $[i]$ is used to denote the $i$-th element in a vector). the model can be trained by optimizing certain loss function that compares $ \bm{y} $ and $ \bm{p} $. 

To capture the fine-grained semantic connections between a document instance and the labels, the document-label similarities are obtained by aggregating the word-label similarities. More specifically, for the text sequence with $ T $ word tokens, \textit{i.e.} $ \bm{x} = [x_1, \dots, x_T] $, the $ i $-th label-aware document representation $ \bm{s}_i = [score(x_1, l_i); \dots; score(x_T, l_i)] $ can be calculate via certain score function. $ \bm{p}_{[i]} $ is then deduced from $ \bm{s}_i $. This process is adapted from the interaction mechanism \cite{du2019explicit}, which is usually used in tasks like natural language inference \cite{wang2016learning}. Based on the idea that labels can be considered as abstraction from their word descriptions, sometimes a label is even a word itself, the word-label similarities can be derived from their embeddings in the latent space by the same way as the word similarity, which is widely studied in word embedding methods such as \textit{GloVe} \cite{pennington2014glove}. 

Note that word embeddings are insufficient to fully represent the meanings of words, especially in the case of \textit{word-sense disambiguation} \cite{navigli2009word}. Take the word "bank" as an example, it has significantly different meanings in the text sequences "go to the bank and change some money" and "flowers generally grow on the river bank", which will cause a variance when matching with labels "economy" and "environment". In order to capture the real semantics of each word, we introduce \textit{RNN}-based word encoder which can take the contextual information of text sequences into consideration.

\subsection{The Poincaré Ball}

In \textit{HyperIM}, both document text and labels are embedded in the hyperbolic space. The hyperbolic space is a homogeneous space that has a constant negative sectional curvature, while the Euclidean space has zero curvature. The hyperbolic space can be described via \textit{Riemannian geometry} \cite{Hopper2011ricci}. Following previous works \cite{nickel2017poincare,ganea2018hyperbolicnn,tifrea2019poincar}, we adopt the \textit{Poincar\'{e} ball}.

An $ n $-dimensional Poincaré ball $ (\mathcal{B}^n, g^{\mathcal{B}^n}) $ is a subset of $ \mathbb{R}^n $ defined by the \textit{Riemannian manifold} $ \mathcal{B}^n = \{\bm{x} \in \mathbb{R}^n\ |\ \|\bm{x}\| < 1\} $ equipped with the \textit{Riemannian metric} $ g^{\mathcal{B}^n} $, where $ \|\cdot\| $ denotes the Euclidean $ L^2 $ norm. As the Poincaré ball is conformal to the Euclidean space \cite{cannon1997hyperbolic}, the Riemannian metric can be written as  $ g^{\mathcal{B}^n}_{\bm{p}} = \lambda_{\bm{p}}^2 g_{\bm{p}}^{\mathbb{R}^n}$ with the conformal factor $ \lambda_{\bm{p}} := \frac{2}{1 - \|\bm{p}\|^2} $ for all $ \bm{p} \in \mathcal{B}^n $, where $ g_{\bm{p}}^{\mathbb{R}^n} = \bm{I}_n $ is the Euclidean metric tensor. It is known that the \textit{geodesic distance} between two points  $ \bm{u}, \bm{v} \in \mathcal{B}^n $ can be induced using the ambient Euclidean geometry as $ d_{\mathcal{B}^n}(\bm{u}, \bm{v}) = \cosh^{-1} (1 + \frac{1}{2} \lambda_{\bm{u}} \lambda_{\bm{v}} \|\bm{u} - \bm{v}\|^2 ) $. This formula demonstrates that the distance changes smoothly \textit{w.r.t.} $ \|\bm{u}\| $ and $ \|\bm{v}\| $, which is key to learn continuous embeddings for hierarchical structures.

With the purpose of generalizing operations for neural networks in the Poincaré ball, the formalism of the M\"{o}bius gyrovector space is used \cite{ganea2018hyperbolicnn}. The M\"{o}bius addition for $ \bm{u}, \bm{v} \in  \mathcal{B}^n $ is defined as $ \bm{u} \oplus \bm{v} = \frac{(1 + 2\langle\bm{u}, \bm{v}\rangle + \|\bm{v}\|^2)\bm{u} + (1 - \|\bm{u}\|^2)\bm{v}}{1 + 2\langle\bm{u}, \bm{v}\rangle + \|\bm{u}\|^2 \|\bm{v}\|^2} $, where $ \langle \cdot, \cdot \rangle $ denotes the Euclidean inner product. The M\"{o}bius addition operation in the Poincaré disk $ \mathcal{B}^2 $ (2-dimensional Poincaré ball) can be visualized in Figure \ref{fig:mob_add}. Then the Poincar\'{e} distance can be rewritten as 
\begin{equation} \label{eq:poincare_dis_mob}
  d_{\mathcal{B}^n}(\bm{u}, \bm{v}) = 2 \tanh^{-1}(\|-\bm{u} \oplus \bm{v} \|).
\end{equation}

The M\"{o}bius matrix-vector multiplication for $ \bm{M} \in \mathbb{R}^{m \times n} $ and $ \bm{p} \in \mathcal{B}^n $ when $ \bm{M}\bm{p} \neq \bm{0} $ is defined as $ \bm{M} \otimes \bm{p} = \tanh( \frac{\|\bm{M}\bm{p}\|}{\|\bm{p}\|} \tanh^{-1}(\|\bm{p}\|)) \frac{\bm{M}\bm{p}}{\|\bm{M}\bm{p}\|}, $
and $ \bm{M} \otimes \bm{p} = \bm{0} $ when $ \bm{M}\bm{p} = \bm{0} $. Moreover, the closed-form derivations of the exponential map $ \exp_{\bm{p}}: T_{\bm{p}} \mathcal{B}^n \to \mathcal{B}^n $ and the logarithmic map $ \log_{\bm{p}}: \mathcal{B}^n \to T_{\bm{p}} \mathcal{B}^n $ for 
$ \bm{p} \in \mathcal{B}^n $, $ \bm{w} \in T_{\bm{p}} \mathcal{B}^n \setminus \{\bm{0}\} $, $ \bm{u} \in \mathcal{B}^n \setminus \{\bm{p}\} $ are given as $ \exp_{\bm{p}}(\bm{w}) = \bm{p} \oplus \bigg( \tanh (\frac{\lambda_{\bm{p}}}{2} \|\bm{w}\|) \frac{\bm{w}}{\|\bm{w}\|} \bigg) $ and $ \log_{\bm{p}}(\bm{u}) = \frac{2}{\lambda_{\bm{p}}} \tanh^{-1}(\|-\bm{p} \oplus \bm{u}\|) \frac{-\bm{p} \oplus \bm{u}}{\|-\bm{p} \oplus \bm{u}\|} $.

These operations make hyperbolic neural networks available \cite{ganea2018hyperbolicnn} and gradient-based optimizations can be performed to estimate the model parameters in the Poincaré ball \cite{becigneul2019riemannian}.

\begin{figure*}
  \centering
  \includegraphics[width=.75\textwidth]{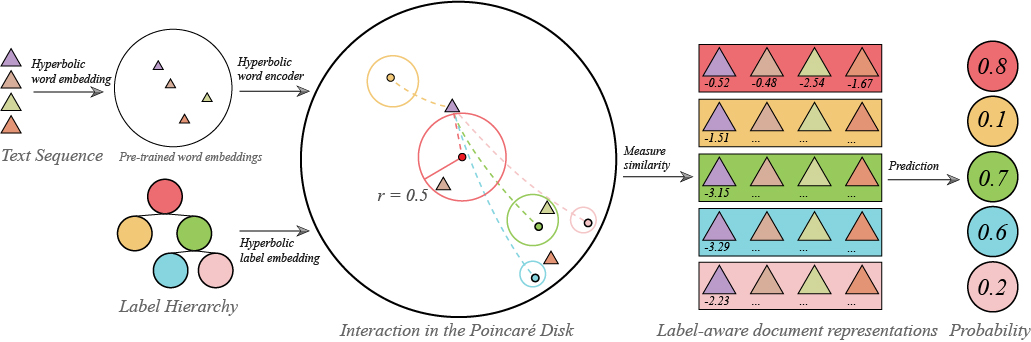}
  \caption{Framework of the Hyperbolic Interaction Model (\textit{HyperIM}). Word-label similarities are measured in the Poincaré Disk. The label nodes are the centres of the hyperbolic circles, which have the same radius. The dash lines are the geodesics from the label nodes to a word node. Note that the hyperbolic centers of the circles in general don't correspond to the Euclidean ones. Labels have the same similarity scores for words embedded on the boundary of their circles.}  
  \label{fig:interaction}
\end{figure*}

\section{Hyperbolic Interaction Model}
\label{sec3}

We design a Hyperbolic Interaction Model (\textit{HyperIM}) for hierarchical multi-label text classification. Given the text sequence of a document, \textit{HyperIM} measures the word-label similarities by calculating the geodesic distance between the jointly embedded words and labels in the Poincaré ball. The word-label similarity scores are then aggregated to estimate the label-aware document representations and further predict the probability for each label. Figure \ref{fig:interaction} demonstrates the framework of \textit{HyperIM}.

\subsection{Hyperbolic Label Embedding}
\label{sec3-label_embed}

The tree-likeness of the hyperbolic space \cite{hamann2018tree} makes it natural to embed hierarchical structures. For instance, Figure \ref{fig:embed_tree} presents a tree embedded in the Poincaré disk, where the root is placed at the origin and the leaves are close to the boundary. It has been shown that any finite tree can be embedded with arbitrary low distortion into the Poincare ball while the distances are approximately preserved \cite{sarkar2011low}. Conversely, it's difficult to perform such embedding in the Euclidean space even with unbounded dimensionality \cite{sala2018representation}. Since the label hierarchy is defined in the set $ \mathcal{T} = \{(l_p, l_q) \ | \ l_p \succeq l_q, \ \  l_p, l_q \in \mathcal{L} \} $, the goal is to maximize the distance between labels without parent-child relation \cite{nickel2017poincare}. Let $ \bm{\Theta}^L = \{\bm{\theta}_i^l\}_{i=1}^C, \ \bm{\theta}_i^l \in \mathcal{B}^k $ be the label embedding set, using Riemannian adaptive optimization methods \cite{becigneul2019riemannian}, $ \bm{\Theta}^L $ can be efficiently estimated by minimizing the loss function
\begin{equation} \label{eq:hier_loss}
  \resizebox{.9\columnwidth}{!}{
    $ \mathcal{L}_{loss}^h = -\sum\limits_{(l_p, \ l_q) \in \mathcal{T}} log \frac{exp \big( -d_{\mathcal{B}^k}(\bm{\theta}_p^l, \ \bm{\theta}_q^l) \big) }{\sum\limits_{l_{q'} \in \mathcal{N}(l_p)} exp \big(-d_{\mathcal{B}^k}(\bm{\theta}_p^l, \ \bm{\theta}_{q'}^l) \big)}$,
  }
\end{equation}
where $ \mathcal{N}(l_p) = \{l_{q'} \vert (l_p, \ l_{q'}) \notin \mathcal{T} \} \cup \{l_p\}$ is the set of negative samples. The obtained $ \bm{\Theta}^L $ can capture the hierarchical structure among labels.

\subsection{Hyperbolic Word Embedding}
\label{sec3-word_embed}

For natural language processing, word embeddings are essential in neural networks as intermediate features. Given the statistics of word co-occurrences in the corpus, we adopt the \textit{Poincaré GloVe} \cite{tifrea2019poincar} to capture the elementary relations between words by embedding them in the hyperbolic space. Let $ X_{ij} $ indicate the times that word $ i $ and word $ j $ co-occur in the same context window, $ \bm{\theta}_i^e \in \mathcal{B}^k $ be the target embedding vector in the $ k $-dimensional Poincaré ball for word $ i $, and $ \tilde{\bm{\theta}}_j^e \in \mathcal{B}^k $ be the context embedding vector for word $ j $. With the aid of Riemannian adaptive optimization methods, the embeddings $ \bm{\Theta}^E = \{ \bm{\theta}_i^e \}_{i=1}^V $ and $ \tilde{\bm{\Theta}}^E = \{ \tilde{\bm{\theta}}_j^e \}_{j=1}^V $ for the corpus with vocabulary size $ V $ are estimated by minimizing the loss function 
\begin{equation}
  \resizebox{.9\columnwidth}{!}{
    $ \mathcal{L}_{loss}^e = \sum\limits_{i, j = 1}^V f(X_{ij}) \big( -h(d_{\mathcal{B}^k} (\bm{\theta}_i^e, \ \tilde{\bm{\theta}}_j^e)) + b_i + \tilde{b}_j - log(X_{ij}) \big)^2 $, 
  }
\end{equation}
where $  b_i $, $ \tilde{b}_j $ are the biases, and the two suggested weight functions are defined as $ f(x) = \min(1, \ (x/100)^{3/4}) $, $ h(x) = cosh^2(x) $. 

Since the \textit{WordNet} hypernym \cite{miller1995wordnet} set $ \mathcal{T}^w = \{(x_p, x_q) \ | \ x_p \succeq x_q \} $, where word $ x_p $ is the hypernym of word $ x_q $ in the corpus, is similar to the label hierarchy, providing the hypernym information to the word embeddings, latent correlations between the two hierarchies can be later captured via interaction. Considering the learned \textit{Poincaré GloVe} embeddings $ \bm{\Theta}^E $ don't explicitly capture the conceptual relations among words, a post-processing step similar to Eq. (\ref{eq:hier_loss}) on top of $ \bm{\Theta}^E $ is further conducted by using Riemannian adaptive optimization methods to minimize the loss function
\begin{equation}
  \resizebox{0.9\columnwidth}{!}{
    $ \mathcal{L}_{loss}^h = -\sum\limits_{(x_p, \ x_q) \in \mathcal{T}} log \frac{exp \big( -d_{\mathcal{B}^k}(\bm{\theta}_p^e, \ \bm{\theta}_q^e) \big) }{\sum\limits_{x_{q'} \in \mathcal{N}(x_p)} exp \big(-d_{\mathcal{B}^k}(\bm{\theta}_p^e, \ \bm{\theta}_{q'}^e) \big)} $,
  }
\end{equation}
where $ \mathcal{N}(x_p) = \{x_{q'} \vert (x_p, \ x_{q'}) \notin \mathcal{T}^w \} \cup \{x_p\}$ is the set of negative samples.

\subsection{Hyperbolic Word Encoder}
\label{sec3-word_encoder}

Considering the \textit{word-sense disambiguation} \cite{navigli2009word}, meanings of polysemous words are difficult to distinguish if the word and label embeddings interact with each other directly, since all the meanings of a word are embedded on the same position. However, polysemous words can usually be inferred from the context.

Given the text sequence of a document with $ T $ word tokens $ \bm{x} = [x_1, \dots, x_T] $, pre-trained hyperbolic word embeddings $ \bm{\Theta}^{E} $ can be used to learn the final word representations according to the text sequence. To consider the sequentiality of the text sequence, we take advantage of the hyperbolic space adaptive \textit{RNN}-based architectures \cite{ganea2018hyperbolicnn}. More specifically, given $ \bm{\Theta}^{e} = [\bm{\theta}^{e}_1, \dots, \bm{\theta}^{e}_T] $ where $ \bm{\theta}^{e}_t \in \bm{\Theta}^E ( t = 1, \dots, T ) $, the hyperbolic word encoder based on the \textit{GRU} architecture adjusts the embedding for each word to fit its context via
\begin{equation} \label{eq:hyper_gru}
  \begin{split}
    & \bm{r}_t = \sigma \big( \log_{\bm{0}} (\bm{W}^r \otimes \bm{\theta}_{t - 1}^w \oplus \bm{U}^r \otimes \bm{\theta}_t^e \oplus \bm{b}^r) \big), \\
    & \bm{z}_t = \sigma \big( \log_{\bm{0}} (\bm{W}^z \otimes \bm{\theta}_{t - 1}^w \oplus \bm{U}^z \otimes \bm{\theta}_t^e \oplus \bm{b}^z) \big), \\
    & \tilde{\bm{\theta}}_t^w = \varphi ( (\bm{W}^g diag(\bm{r}_t)) \otimes \bm{\theta}_{t - 1}^w \oplus \bm{U}^g \otimes \bm{\theta}_t^e \oplus \bm{b}^g ), \\
    & \bm{\theta}_t^w = \bm{\theta}_{t - 1}^w \oplus diag(\bm{z}_t) \otimes (-\bm{\theta}_{t - 1}^w \oplus \tilde{\bm{\theta}}_t^w),
  \end{split}
\end{equation}
where $ \bm{\Theta}^w = [\bm{\theta}_1^w, \dots, \bm{\theta}_T^w] $ denotes the encoded embeddings for the text sequence, the initial hidden state $ \bm{\theta}_0^w\ := \bm{0} $, $ \bm{r}_t $ is the reset gate, $ \bm{z}_t $ is the update gate, $ diag(\cdot) $ denotes the diagonal matrix with each element in the vector on its diagonal, $ \sigma $ is the \textit{sigmoid} function, $ \varphi $ is a pointwise non-linearity, typically \textit{sigmoid}, \textit{tanh} or \textit{ReLU}. Since the hyperbolic space naturally has non-linearity, $ \varphi $ can be \textit{identity} (no non-linearity) here. 

The formula of the \textit{hyperbolic GRU} is derived by connecting the M\"{o}bius gyrovector space with the Poincaré ball \cite{ganea2018hyperbolicnn}. The six weights $ \bm{W}, \bm{U} \in \mathbb{R}^{k \times k} $ are trainable parameters in the Euclidean space and the three biases $ \bm{b} \in \mathcal{B}^k $ are trainable parameters in the hyperbolic space (the superscripts are omitted for simplicity). Thus the weights $ \bm{W} $ and $ \bm{U} $ are updated via vanilla optimization methods, and the biases $ \bm{b} $ are updated with Riemannian adaptive optimization methods. $ \bm{\Theta}^w $ will be used for measuring the word-label similarities during the following interaction process.

\subsection{Interaction in the Hyperbolic Space}
\label{sec3-interaction}

The major objective of text classification is to build connections from the word space to the label space. In order to capture the fine-grained semantic information, we first construct the label-aware document representations, and then learn the mappings between the document instance and the labels.

\subsubsection{Label-aware Document Representations}
Once the encoded word embeddings $ \bm{\Theta}^w $ and label embeddings $ \bm{\Theta}^L $ are obtained, it's expected that every pair of word and label embedded close to each other based on their geodesic distance if they are semantically similar. Note that cosine similarity \cite{wang2017bilateral} is not appropriate to be the metric since there doesn't exist a clear hyperbolic inner-product for the the Poincar\'{e} ball \cite{tifrea2019poincar}, so the geodesic distance is more intuitively suitable. The similarity between the $ t $-th word $ x_t (t = 1, \dots, T) $ and the $ i $-th label $ l_i (i = 1, \dots, C) $ is calculated as $ score(x_t, l_i) = - d_{\mathcal{B}^k}(\bm{\theta}_t^w, \bm{\theta}_i^l) $, where $ \bm{\theta}_t^w $ and $ \bm{\theta}_i^l $ are their corresponding embeddings, $ d_{\mathcal{B}^k}(\cdot, \cdot) $ is the Poincaré distance function defined in Eq. (\ref{eq:poincare_dis_mob}). The $ i $-th label-aware document representation can be formed as the concatenation of all the similarities along the text sequence, \textit{i.e.} $ \bm{s}_i = [score(x_1, l_i); \dots; score(x_T, l_i)] $. The set $ \mathcal{S} = \{ \bm{s}_i \}_{i = 1}^C $ acquired along the labels can be taken as the label-aware document representations under the hyperbolic word and label embeddings. 

\subsubsection{Prediction}
Given the document representations in $ \mathcal{S} $, predictions can be made by a fully-connected layer and an output layer. The probability of each label for the document instance can be obtained by
\begin{equation} \label{eq:mlp_aggre}
  p_{i} = \sigma(\bm{W}^e \varphi(\bm{W}^f \bm{s}_i)), \forall \bm{s}_i \in \mathcal{S}, i = 1, \dots, C,
\end{equation}
where $ \sigma $ is the \textit{sigmoid} function, $ \varphi $ is a non-linearity. The weights $ \bm{W}^e \in \mathbb{R}^{1 \times (T/2)} $ and $ \bm{W}^f \in \mathbb{R}^{(T/2) \times T} $ are trainable parameters. 

\subsection{Partial Interaction}

During the above interaction process, the amount of computation increases with the number of labels. When the output label space is large, it's a burden to calculate the label-aware document representations. On account of the fact that only a few labels are assigned to one document instance, we propose to use a negative sampling method to improve the scalability during training. Let $ \mathcal{L}^+ $ denote the set of true labels and $ \mathcal{L}^- $ denote the set of randomly selected negative labels, the model is trained by minimizing the loss function which is derived from the \textit{binary cross-entropy} loss as it is commonly used for \textit{MLC} \cite{liu2017deep}, \textit{i.e.} 
\begin{equation} \label{eq:bceloss}
  \mathcal{L}_{loss}^b = - \big( \sum_{i \in \mathcal{L}^+ } log(p_{i}) + \sum_{j \in \mathcal{L}^- } log(1 - p_{j}) \big).
\end{equation}

The hyperbolic parameters, \textit{i.e.} $ \bm{\Theta}^E, \ \bm{\Theta}^L $ and $ \bm{b} $ in the hyperbolic word encoder, are updated via Riemannian adaptive optimization methods. The Euclidean parameters, \textit{i.e.} $ \bm{W}, \bm{U} $ in the hyperbolic word encoder and  $ \bm{W} $ in the prediction layers, are updated via vanilla optimization methods. \textit{Partial interaction} can significantly reduce the memory usage during training especially when the label set is large.

\section{Experiments}
\label{sec4}

\subsubsection{Datasets}
Experiments are carried out on three publicly available multi-label text classification datasets, including the small-scale \textit{RCV1} \cite{lewis2004rcv1}, the middle-scale \textit{Zhihu}\footnote{\url{https://biendata.com/competition/zhihu/}.} and the large-scale \textit{WikiLSHTC} \cite{ioannis2015lshtc}. All the datasets are equipped with labels that explicitly exhibit a hierarchical structure. Their statistics can be found in Table \ref{table:dataset}.

\subsubsection{Pre-processing}
All words are converted to lower case and padding is used to handle the various lengths of the text sequences. Different maximum lengths are set for each dataset according to the average number of words per document in the training set, \textit{i.e.} 300 for \textit{RCV1}, 50 for \textit{Zhihu} and 150 for \textit{WikiLSHTC}.

\begin{table}
  \caption[Caption for LOF]{Statistics of the datasets: $ N_{train} $ and $ N_{test} $ are the number of training and test instances, $ L $ is the number of labels, $ \hat{L} $ is the average number of label per document, $ \tilde{L} $ is the average number of documents per label, $ W_{train} $ and $ W_{test} $ denote the average number of words per document in the training and test set respectively.}
  \label{table:dataset}
  \centering
  \resizebox{\columnwidth}{!}{\begin{tabular}{lcccccccc}
    \toprule
    Dataset & $ N_{train} $ & $ N_{test} $  & $ L $ & $ \hat{L} $ & $ \tilde{L} $ & $ W_{train} $ & $ W_{test} $  \\
    \midrule
    \textit{RCV1}  & 23,149    & 781,265 & 103    & 3.18 & 729.67  & 259.47  & 269.23 \\
    \textit{Zhihu} & 2,699,969 & 299,997 & 1,999  & 2.32 & 3513.17 & 38.14   & 35.56  \\
    \textit{WikiLSHTC} & 456,886   & 81,262  & 36,504 & 1.86 & 4.33    & 117.98  & 118.31 \\
    \bottomrule
  \end{tabular}}
\end{table}

\begin{table*}
  \caption{Results in \textit{P@k} and \textit{nDCG@k}, bold face indicates the best in each line.}
  \label{table:result}
  \centering
  \resizebox{.7\textwidth}{!}{\begin{tabular}{llcccccccc}
    \toprule
    Dataset & Metrics & \ \ \textit{EXAM}\ \  & \textit{SLEEC} & \ \ \textit{DXML}\ \  & \textit{HR-DGCNN} & \textit{HMCN-F} & \textit{HyperIM} \\
    \midrule
          & $ \textit{P@1} $          & 95.98 & 94.45 & 95.27 & 95.17 & 95.35 & \textbf{96.78}  \\
          & $ \textit{P@3} $          & 80.83 & 78.60 & 77.86 & 80.32 & 78.95 & \textbf{81.46} \\
    \textit{RCV1} & $ \textit{P@5} $  & 55.80 & 54.24 & 53.44 & 55.38 & 55.90 & \textbf{56.79} \\
          & $ \textit{nDCG@3} $       & 90.74 & 90.05 & 89.69 & 90.02 & 90.14 & \textbf{91.52} \\
          & $ \textit{nDCG@5} $       & 91.26 & 90.32 & 90.24 & 90.28 & 90.82 & \textbf{91.89} \\
    \midrule
          & $ \textit{P@1} $          & 51.41 & 51.34 & 50.34 & 50.97 & 50.24 & \textbf{52.14} \\
          & $ \textit{P@3} $          & 32.81 & 32.56 & 31.21 & 32.41 & 32.18 & \textbf{33.66} \\
    \textit{Zhihu} & $ \textit{P@5} $ & 24.29 & 24.23 & 23.36 & 23.87 & 24.09 & \textbf{24.99} \\
          & $ \textit{nDCG@3} $       & 49.32 & 49.27 & 47.92 & 49.02 & 48.36 & \textbf{50.13} \\
          & $ \textit{nDCG@5} $       & 50.74 & 49.71 & 48.65 & 49.91 & 49.21 & \textbf{51.05} \\
    \midrule
              & $ \textit{P@1} $          & 54.90 & 53.57 & 52.02 & 52.67 & 53.23 & \textbf{55.06} \\
              & $ \textit{P@3} $          & 30.50 & 31.25 & 30.57 & 30.13 & 29.32 & \textbf{31.73} \\
    \textit{WikiLSHTC} & $ \textit{P@5} $ & 22.02 & 22.46 & 21.66 & 22.85 & 21.79 & \textbf{23.08} \\
              & $ \textit{nDCG@3} $       & 49.50 & 46.06 & 47.97 & 49.24 & 48.93 & \textbf{50.46} \\
              & $ \textit{nDCG@5} $       & 50.46 & 47.52 & 48.14 & 50.42 & 49.87 & \textbf{51.36} \\
    \bottomrule
  \end{tabular}}
\end{table*}

\subsubsection{Evaluation metrics}
We use the rank-based evaluation metrics which have been widely adopted for multi-label classification tasks, \textit{i.e.} \textit{Precision@k} (\textit{P@k} for short) and \textit{nDCG@k} for $ k = 1, 3, 5 $ \cite{bhatia2015sparse,liu2017deep,zhang2018deep}. Let $ \bm{y} \in \{0, 1\}^C $ be the ground truth label vector for a document instance and $ \bm{p} \in [0, 1]^C $ be the predicted label probability vector. \textit{P@k} records the fraction of correct predictions in the top $ k $ possible labels. Let the vector $ \bm{r} \in \{1, \dots, C\}^k $ denote the indices for $ k $ most possible labels in descending order, \textit{i.e.} the $ \bm{r}_{[1]} $-th label has the largest probability to be true, then the metrics are defined as
\begin{equation} \label{eq:metrics}
  \begin{split}
    & \textit{P@k} = \frac{1}{k} \sum_{i = 1}^k \bm{y}_{[\bm{r}_{[i]}]}, \\
    & \textit{nDCG@k} =  \frac{\sum_{i = 1}^k \bm{y}_{[\bm{r}_{[i]}]} / log(i + 1)}{\sum_{i = 1}^{{\min(k,\ \| \bm{y} \|_0)}} 1 / log(i + 1)},  
  \end{split}
\end{equation}
where  $ \| \bm{y} \|_0 $ denotes the number of true labels, \textit{i.e.} the number of $ 1 $ in $ \bm{y} $. The final results are averaged over all the test document instances. Notice that \textit{nDCG@1} is omitted in the results since it gives the same value as \textit{P@1}.

\subsubsection{Baselines}
To demonstrate the effectiveness of \textit{HyperIM} on the benchmark datasets, five comparative multi-label classification methods are chosen. \textit{EXAM} \cite{du2019explicit} is the state-of-the-art interaction model for text classification. \textit{EXAM} use pre-trained word embeddings in the Euclidean space, its label embeddings are randomly initialized. To calculate the similarity scores, \textit{EXAM} uses the dot-product between word and label embeddings. \textit{SLEEC} \cite{bhatia2015sparse} and \textit{DXML} \cite{zhang2018deep} are two label-embedding methods. \textit{SLEEC} projects labels into low-dimensional vectors which can capture label correlations by preserving the pairwise distance between them. \textit{SLEEC} uses the \textit{k-nearest neighbors} when predicting, and clustering is used to speed up its prediction. Ensemble method is also used to improve the performance of \textit{SLEEC}. \textit{DXML} uses \textit{DeepWalk} \cite{perozzi2014deepwalk} to embed the label co-occurrence graph into vectors, and uses neural networks to map the features into the embedding space. \textit{HR-DGCNN} \cite{peng2018deepgraphcnn} and \textit{HMCN-F} \cite{wehrmann2018hierarchical} are two neural network models specifically designed for hierarchical classification tasks. Taking advantage of the label hierarchy, \textit{HR-DGCNN} adds a regularization term on the weights of the fully-connected layer. The original \textit{HMCN-F} can't take in the raw text data. To make \textit{HMCN-F} more competitive, \textit{CNN}-based architecture similar to \textit{XML-CNN} \cite{liu2017deep} is adopted to extract the primary features. \textit{HMCN-F} then fits its neural network layers to the label hierarchy, each layer focuses on predicting the labels in the corresponding hierarchical level.

\subsubsection{Hyperparameters}
To evaluate the baselines, hyperparameters recommended by their authors are used. \textit{EXAM} uses the label embeddings dimension 1024 for \textit{RCV1} and \textit{Zhihu}, on account of the scalability, it is set to 300 for \textit{WikiLSHTC}. The word embedding dimension of \textit{EXAM} is set to 300 for \textit{RCV1} and \textit{WikiLSHTC}, 256 for \textit{Zhihu}.  \textit{SLEEC} uses embedding dimension 100 for \textit{RCV1}, 50 for \textit{Zhihu} and \textit{WikiLSHTC}. \textit{DXML} uses embedding dimension 100 for \textit{RCV1}, 300 for \textit{Zhihu} and \textit{WikiLSHTC}. The word embedding dimension of \textit{HR-DGCNN} is 50 and its window size is set to be 5 to construct the graph of embeddings. Note that the original \textit{HMCN-F} can't take in the raw text data and further use the word embeddings. To make \textit{HMCN-F} more competitive, \textit{HMCN-F} uses \textit{CNN}-based architecture similar to \textit{XML-CNN} \cite{liu2017deep} to extract the primary features from the raw text, with the word embedding dimension set to 300 for \textit{RCV1} and \textit{WikiLSHTC}, 256 for \textit{Zhihu}. 

\subsubsection{Experimental details}
The hyperparameters used to pre-train the \textit{Poincaré GloVe} and the vanilla \textit{GloVe} are the same values recommended by the authors. \textit{Partial routing} is not applied when training \textit{HyperIM} for \textit{RCV1}, since the label set is not large. When evaluating \textit{HyperIM} for \textit{Wi\textit{kiLSHTC}}, due to the scalability issue, the label set is divided into serval groups. Models shared the same word embeddings, label embeddings and word encoder parameters predict different groups accordingly. This is feasible since the parameters in the prediction layers are the same for all the labels.

On account of the numeric error issue caused by the constrain $ \|\bm{p}\| < 1 $ for $ \bm{p} \in \mathcal{B}^k $ when the embedding dimension $ k $ is large, the workaround is taken to address this issue, \textit{i.e.} the embedding vector is a concatenation of vectors in the low-dimensional Poincaré ball. Consequently, the embedding dimension for \textit{HyperIM} is $ 75 \times 2D $ as it generally outperforms the baselines.

\subsubsection{Numerical Errors}
When the hyperbolic parameters go to the border of the Poincaré ball, gradients for the Möbius operations are not defined. Thus the hyperbolic parameters are always projected back to the ball with a radius $ 1 - 10^{-5} $. Similarly when they get closer to $ \bm{0} $, a small perturbation ($ \epsilon = 10^{-15} $) is applied before they are used in the Möbius operations.

\subsubsection{Optimization}
The Euclidean parameters are updated via \textit{Adam}, and the hyperbolic parameters are updated via Riemannian adaptive \textit{Adam} \cite{becigneul2019riemannian}. The learning rate is set to 0.001, other parameters take the default values. Early stopping is used on the validation set to avoid overfitting.

\subsubsection{Results}
As shown in Table \ref{table:result}, \textit{HyperIM} consistently outperforms all the baselines. \textit{HyperIM} effectively takes advantage of the label hierarchical structure comparing with \textit{EXAM}, \textit{SLEEC} and \textit{DXML}. \textit{EXAM} uses the interaction mechanism to learn word-label similarities, whereas clear connections between the words and the label hierarchy can't be captured since its label embeddings are randomly initialized. The fact that \textit{HyperIM} achieves better results than \textit{EXAM} further confirms that \textit{HyperIM} benefits from the retention of the hierarchical label relations. Meanwhile, the word embeddings learned by \textit{HyperIM} have strong connections to the label structure, which is helpful to the measurement of word-label similarities and the acquirement of the label-aware document representations. \textit{SLEEC} and \textit{DXML} take the label correlations into account. However, the label correlations they use are captured from the label matrix, \textit{e.g.} embedding the label co-occurrence graph, which may be influenced by \textit{tail labels}. For \textit{HyperIM}, the label relations are determined from the label hierarchy, so the embeddings of labels with parent-child relations are dependable to be correlated.

As expected, \textit{HyperIM} is superior to the existing hierarchical classification methods \textit{HR-DGCNN} and \textit{HMCN-F}, even though they take advantage of the label hierarchy information. By investigating the properties of these three methods, we summarize the main reasons as follows. \textit{HR-DGCNN} adds the regularization terms based on the assumption that labels with parent-child relations should have similar weights in the fully-connected layer, which may not always be true in real applications. \textit{HMCN-F} highly depends on the label hierarchy, it assumes that different paths pass through the same number of hierarchical levels. Unfortunately, in the real data, different paths may have totally different lengths. \textit{HyperIM} models the label relations by embedding the label hierarchy in the hyperbolic space. Any hierarchical structure can be suitable and labels aren't required to sit on a specific hierarchical level, which makes \textit{HyperIM} less reliant on the label hierarchy. Furthermore, \textit{HyperIM} can learn the word-label similarities and preserve the label relations simultaneously to acquire label-aware document representations, whereas \textit{HR-DGCNN} and \textit{HMCN-F} treat document words and labels separately.

\subsection{Ablation Test}

\begin{figure*}
  \centering
  \includegraphics[width=.8\textwidth]{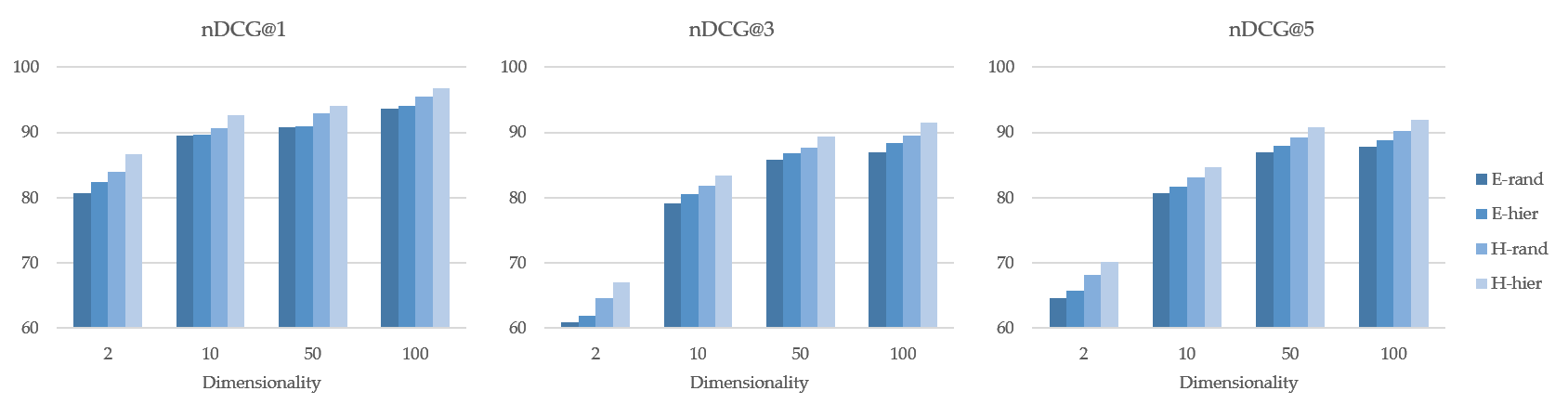}
  \caption{Results in \textit{nDCG@k} for the ablation test. \textit{E-rand} and \textit{H-rand} denote \textit{EuclideanIM} and \textit{HyperIM} take the randomly initialized label embeddings respectively, \textit{E-hier} and \textit{H-hier} take the same label embeddings initialized according to the hierarchical label relations.}  
  \label{fig:ablation}
\end{figure*}

In order to show the characteristics of \textit{HyperIM} and justify the superiority of the hyperbolic space for hierarchical multi-label text classification, we are interested in comparing it with an analogous model in the Euclidean space.

\subsubsection{Euclidean Interaction Model} The analogous model in the Euclidean space (\textit{EuclideanIM}) has a similar architecture as \textit{HyperIM}. \textit{EuclideanIM} takes the vanilla pre-trained \textit{GloVe} word embeddings \cite{pennington2014glove} and uses the vanilla \textit{GRU} \cite{chung2014empirical} as the word encoder. The label embeddings are randomly initialized for \textit{E-rand}, while \textit{E-hier} takes the same label embeddings initialized by the hierarchical label relations as \textit{H-hier}. The word-label similarities are computed as the negative of the Euclidean distance between their embeddings, \textit{i.e.} $score(x_t, l_i) = - \| \bm{\theta}_t^w - \bm{\theta}_i^l \| $ for $ \bm{\theta}_t^w, \bm{\theta}_i^l \in \mathbb{R}^k $. The same architecture of the prediction layers is adopted.

\subsubsection{Results}

Figure \ref{fig:ablation} shows the \textit{nDCG@k} results for different embedding dimensions on the \textit{RCV1} dataset. The fact that \textit{E-hier} slightly outperforms \textit{E-rand} indicates that the label correlations provide useful information for classification. However, \textit{H-rand} still achieves better results than the Euclidean models even without the hierarchical label structure information, which confirms that the hyperbolic space is more suitable for \textit{HMLC}. For \textit{E-hier}, the hierarchical label structure is not appropriate to be embedded in the Euclidean space, thus it can't fully take advantage of such information. \textit{HyperIM} generally outperforms \textit{EuclideanIM} and achieves significant improvement especially in low-dimensional latent space. \textit{H-hier} takes in the label correlations and outperforms \textit{H-rand} as expected.

\subsubsection{Interaction Visualization}

\begin{figure}
  \centering
  \includegraphics[width=1.\columnwidth]{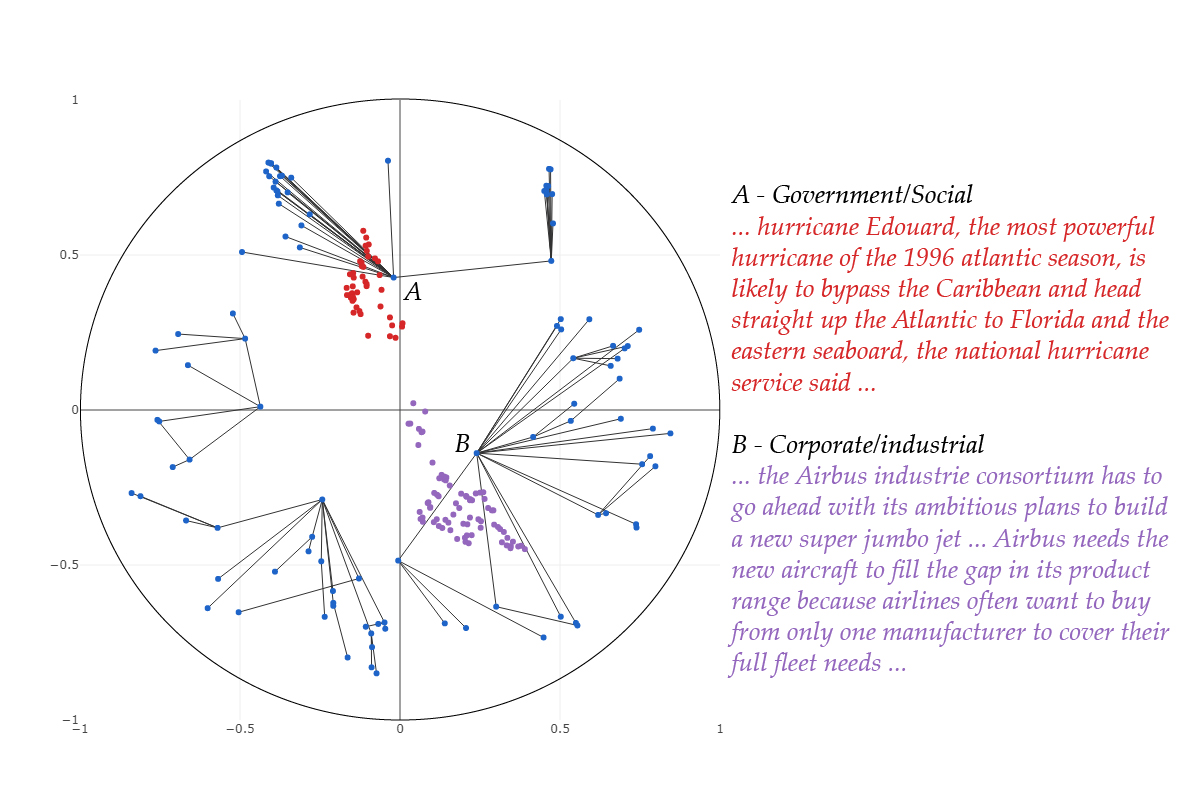}
  \caption{Visualization of labels (blue nodes) and words jointly embedded in the Poincaré disk. The connected labels denote that they are related to each other.}  
  \label{fig:inter_viz}
\end{figure}

The 2-dimensional hyperbolic label embeddings and the encoded word embeddings (not the pre-trained word embeddings) can be visualize jointly in the Poincaré disk as shown in Figure \ref{fig:inter_viz}. The hierarchical label structure which can represent the parent-child relations between labels is well preserved by \textit{HyperIM}. Note that the embedded label hierarchy resembles the embedded tree in Figure \ref{fig:embed_tree}. The top-level nodes (\textit{e.g.} the label node \textit{A}) are embedded near the origin of the Poincaré disk, while the leaf nodes are close to the boundary. The hierarchical label relations are well modeled by such tree-like structure. Moreover, in the dataset, the top-level labels are not connected to an abstract "root". The structure of the embedded label hierarchy still suggests that there should be a "root" that connects all the top-level labels to put at the very origin of the Poincaré disk, which indicates that \textit{HyperIM} can really make use of the hierarchical label relations.

The explicit label correlations can further help \textit{HyperIM} to learn to encode the word embeddings via interaction. The encoded text of a document instance are generally embedded close to the assigned labels. This clear pattern between the encoded word embeddings and the label hierarchy indicates that \textit{HyperIM} learns the word-label similarities with the label correlations taken into consideration. This is the main reason that \textit{HyperIM} outperforms \textit{EuclideanIM} significantly in low dimensions. Some of the words such as "\textit{the}", "\textit{is}" and "\textit{to}" don't provide much information for classification, putting these words near the origin can make them equally similar to labels in the same hierarchical level. A nice by-product is that the predicted probabilities for labels in the same hierarchical level won’t be influenced by these words. Moreover, the variance of word-label distance for labels in different hierarchical levels make parent labels distinguishable from child labels, \textit{e.g.} top-level labels can be made different from the leaf labels since they are generally closer to the word embeddings. Such difference suggests that \textit{HyperIM} treats the document instances differently along the labels in different hierarchical levels.

\subsubsection{Embedding Visualization}
  
\begin{figure*}z
  \begin{subfigure}{.45\linewidth}
    \centering
    \includegraphics[height=5cm]{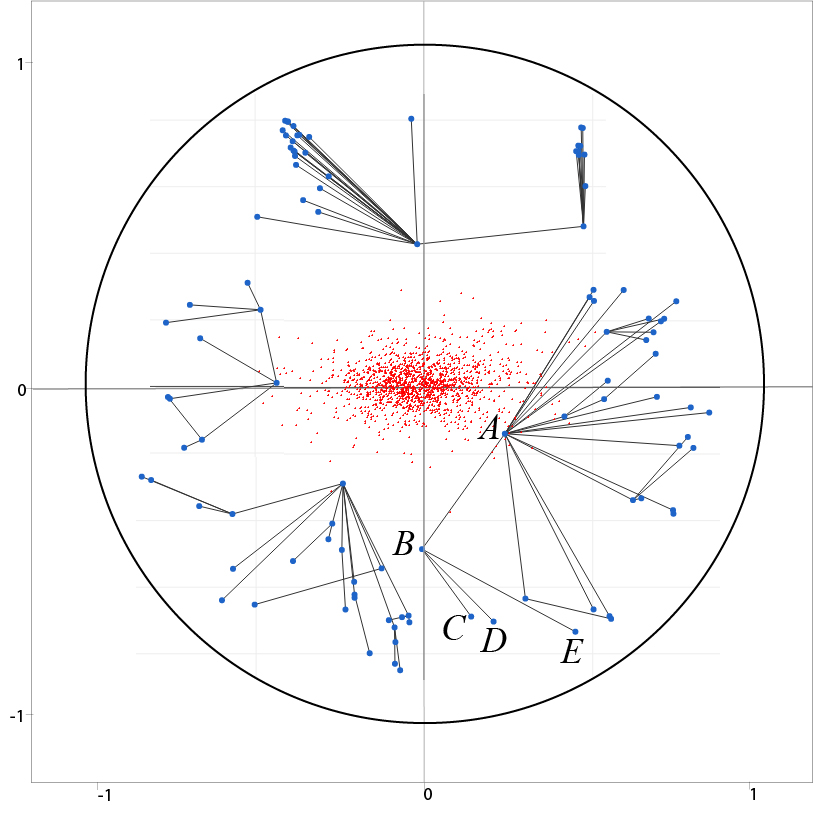}
    \caption{Word/Label embeddings in \textit{2-D} Poincaré disk}
    \label{fig:hyperbolic_trained}
  \end{subfigure}
  \begin{subfigure}{.45\linewidth}
    \centering
    \includegraphics[height=5cm]{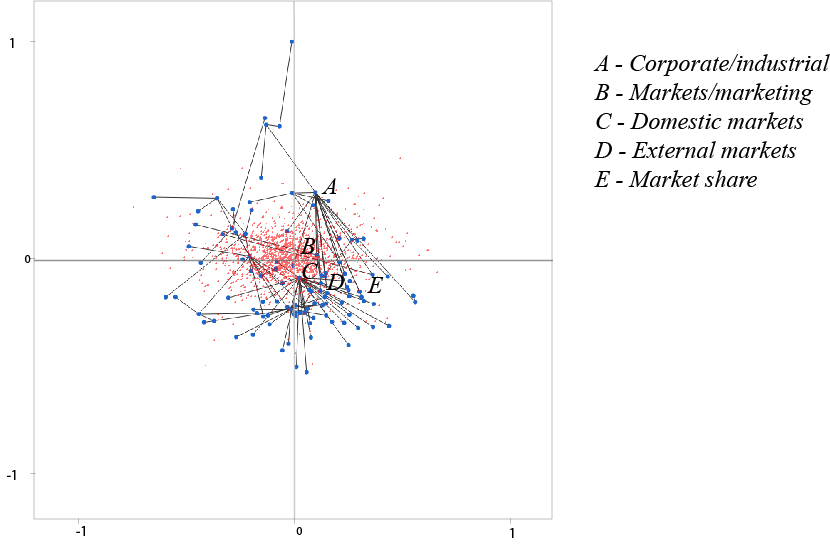}
    \caption{Word/Label embeddings in \textit{2-D} Euclidean space}
    \label{fig:euclidean_trained}
  \end{subfigure}
  \caption{ Words (red points) and labels (blue nodes) from \textit{RCV1} jointly embedded by (a) \textit{H-hier} and (b) \textit{E-hier}. The connected labels denote that they are related to each other.}
  \label{fig:trained_embed}
\end{figure*}

The word/label embedding in the 2-dimensional latent space obtained by \textit{H-hier} and \textit{E-hier} are demonstrated respectively in Figure \ref{fig:trained_embed}. As expected, the hierarchical label structure which can represent the parent-child relations between labels is well preserved by \textit{HyperIM} (as shown in Figure \ref{fig:hyperbolic_trained}), while the label embeddings in the Euclidean space are less interpretable as the paths among labels intersect each other. Consequently, it is hard to find the hierarchical label relations in the Euclidean space. Moreover, the word embeddings are generally near the origin, which gives the word encoder the opportunities to make a piece of text of a document instance tend to different directions where the true labels locate. Whereas it's more difficult to adjust the word embeddings in the Euclidean space since different label hierarchies are not easy to distinguish.

\section{Related Work}
\label{sec5}

\subsection{Hierarchical Multi-Label Classification}

The existing methods dedicating to hierarchical classification usually focus on the design of loss functions or neural network architectures \cite{cerri2015hierarchical}. Traditional hierarchical classification methods optimize a loss function locally or globally \cite{silla2011survey}. Local methods are better at capturing label correlations, whereas global methods are less computationally expensive. Researchers recently try to use a hybrid loss function associated with specifically designed neural networks, \textit{e.g.} \textit{HR-DGCNN} \cite{peng2018deepgraphcnn}. The archetype of \textit{HMCN-F} \cite{wehrmann2018hierarchical} employs a cascade of neural networks, where each neural networks layer corresponds to one level of the label hierarchy. Such neural network architectures generally require all the paths in the label hierarchy to have the same length, which limits their application. Moreover, on account of the fact that labels in high hierarchical levels usually contain much more instances than labels in low levels, whereas neural network layers for low levels need to classify more labels than layers for high levels, such architectures also lead to imbalance classification.

\subsection{Hyperbolic Deep Learning}

Research on representation learning \cite{nickel2017poincare} indicates that the hyperbolic space is more suitable for embedding symbolic data with hierarchical structures than the Euclidean space, since the tree-likeness properties \cite{hamann2018tree} of the hyperbolic space make it efficient to learn hierarchical representations with low distortion \cite{sarkar2011low}. Since linguistic ontologies are innately hierarchies, hierarchies are ubiquitous in natural language, (\textit{e.g.} \textit{WordNet} \cite{miller1995wordnet}). Some works lately demonstrate the superiority of the hyperbolic space for natural language processing tasks such as textual entailment \cite{ganea2018hyperbolicnn}, machine translation \cite{gulcehre2018hyperbolic} and word embedding \cite{tifrea2019poincar}. 

\subsubsection{Riemannian optimization}

In the same way that gradient-based optimization methods are used for trainable parameters in the Euclidean space, the hyperbolic parameters can be updated via Riemannian adaptive optimization methods \cite{becigneul2019riemannian}. For instance, \textit{Riemannian adaptive SGD} updates the parameters $ \bm{\theta} \in \mathcal{B}^k $ by $ \bm{\theta}_{t + 1} = \exp_{\bm{\theta}_t} (-\eta \nabla_R \mathcal{L}(\bm{\theta}_t)) $, where $ \eta $ is the learning rate, and the Riemannian gradient $ \nabla_R \mathcal{L}(\bm{\theta}_t) \in T_{\bm{\theta}} \mathcal{B}^k $ is the rescaled Euclidean gradient, \textit{i.e.} $ \nabla_R \mathcal{L}(\bm{\theta}) = \frac{1}{\lambda_{\bm{\theta}}^2} \nabla_E \mathcal{L}(\bm{\theta}) $ \cite{wilson2018gradient}.

\section{Conclusion}
\label{sec6}

The hierarchical parent-child relations between labels can be well modeled in the hyperbolic space. The proposed \textit{HyperIM} is able to explicitly learn the word-label similarities by embedding the words and labels jointly and preserving the label hierarchy simultaneously. \textit{HyperIM} acquires label-aware document representations to extract the fine-grained text content along each label, which significantly improves the hierarchical multi-label text classification performance. Indeed, \textit{HyperIM} makes use of the label hierarchy, whereas there is usually no such hierarchically organized labels in practice, especially for extreme multi-label classification (\textit{XMLC}). Nevertheless, the labels in \textit{XMLC} usually follow a power-law distribution due to the amount of \textit{tail labels} \cite{babbar2018adversarial}, which can be traced back to hierarchical structures \cite{ravasz2003hierarchical}. Thus, it will be interesting to extend  \textit{HyperIM} for \textit{XMLC} in the future. Our code is publicly available to facilitate other research.

\section{Acknowledgments}
This work was supported in part by the National Natural Science Foundation of China under Grant 61822601, 61773050, and 61632004; the Beijing Natural Science Foundation under Grant Z180006; the Beijing Municipal Science \& Technology Commission under Grant Z181100008918012; National Key Research and Development Program (2017YFC1703506) ;the Fundamental Research Funds for the Central Universities (2019JBZ110).

\clearpage
\small
\bibliographystyle{aaai}
\bibliography{aaai}

\end{document}